\pdfoutput=1
\documentclass{article}
\usepackage{iclr2027_conference,times}
\usepackage{amsmath,amsfonts,bm}
\usepackage{graphicx}
\usepackage[table]{xcolor}
\usepackage{booktabs}
\usepackage{amsmath}
\usepackage{amssymb}
\usepackage{capt-of}
\usepackage{algorithm}
\usepackage{xspace}
\usepackage{algpseudocode}
\usepackage{hyperref}
\makeatletter
\DeclareRobustCommand\onedot{\futurelet\@let@token\@onedot}
\def\@onedot{\ifx\@let@token.\else.\null\fi\xspace}
\providecommand{\ie}{}\renewcommand{\ie}{\emph{i.e}\onedot}
\makeatother

\newcommand{\method}{iSEE}
\newcommand{\privrow}{\rowcolor{gray!14}}
\newcommand{\priv}[1]{\textcolor{black!55}{#1}}
\newcommand{\ten}{TEN}

\iclrfinalcopy

\title{\method{}: Object Permanence through Self-Supervision}
\author{Pramish Paudel, Ajad Chhatkuli, Luc Van Gool \& Danda Pani Paudel \\
INSAIT, Sofia University ``St.\ Kliment Ohridski''}

\begin{document}

\maketitle
\lhead{Preprint}

\begin{center}
\vspace{-0.8em}
\includegraphics[width=\linewidth]{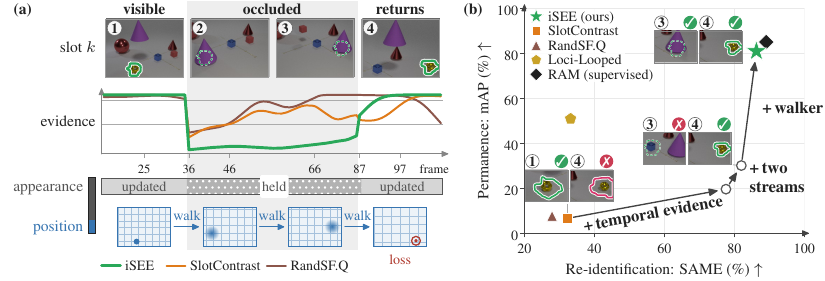}
\vspace{-1.4em}
\captionof{figure}{\textbf{\method{} achieves self-supervised object permanence, close to a supervised tracker.}
(a) Slot $k$ (green) holds the ball; dashed, its mask predicted with \method{} during the occlusion.
A slot's evidence is the attention it receives from the frame's patches: \method{}'s clearly signals
the occlusion and the reappearance. \method{}'s slot is separated into appearance and position; during the occlusion the appearance is held and the
position keeps walking, learnt only from where it reappears (red ring). (b) On LA-CATER (static), each
arrow adds one of our contributions to the existing baseline, SlotContrast~\citep{slotcontrast}.}
\label{fig:teaser}
\end{center}

\begin{abstract}
Object permanence, keeping track of an object's identity and position while it is occluded, is central to video
representations that track, predict and plan. Trackers that achieve it learn from boxes, track identities and visibility labels. On the other hand, self-supervised object-centric
methods discover objects without labels: through slot attention, it represents a video as
slots that bind to objects and follow them across frames. However, these slots are lost under occlusion, making the desired permanence impossible. Reasoning permanence
is a hard problem because it requires to detect when an object becomes occluded, re-identify when object reappears, and keep
the object's hidden position continuous, using reapperance as the only learning cue. To address this, we propose \method{}, a novel
framework that offers all three aforementioned requirements, without any labels whatsoever. We built \method{} using the following three proposed components: (i) Object evidence modelling: a slot's attention, compared with
its own past, reveals when its object is hidden. (ii) Appearance-position separation: two
slot streams let the appearance be held for re-identification while the position keeps changing. (iii) Permanence
from reappearance: a walker follows the hidden object's position, trained only on where the object reappears. On
\mbox{LA-CATER} static, \method{} returns a reappearing object to its own slot after 86\,\% of occlusions, against
32\,\% for SlotContrast~\citep{slotcontrast}, and localises it while hidden within 4.1~mAP of the label-trained SoTA RAM~\citep{ram}. The two streams also
allow downstream planning, with the position stream as the action of a world model.
\end{abstract}

\section{Introduction}

We are interested in occlusion-robustness and object permanence in self-supervised object-centric video representations~\citep{slotcontrast,savi,steve,videosaur} obtained through slot attention~\citep{slotattention}. These models represent a video as a set of evolving slots, \ie, set of vectors
each bound to an object or a region of the frame, and updated from frame
to frame. Thus, downstream reasoning and predictions can use object representation with identity, rather than pixels or patches.
Specifically, SlotContrast uses contrastive learning on slot evolution, showing improvement in representation over its predecessors. However, slot representations trained thus, without additional labels, are occlusion sensitive~\citep{videosaur,slotcontrast}. SlotContrast regards \emph{``consistency during long-term full
occlusions"} as {an open problem} \citep{slotcontrast}.

The problem stems from the implicit assumption that objects are always visible. This justifies the existing methods to update the object slots in every frame. Consequently, under occlusions, the hidden object's slot can simply change its identity to that of a visible patch. Intuitively, under occlusions, object appearance must not be updated, while its position might still change. This relates directly to the second issue that a single slot vector represents both an object's appearance and position. This can lead to the object representation to be biased by position, disregarding visible evidence or lack thereof, under occlusions. Previous approaches to learning object permanence rely on extra supervision: boxes of visible objects~\citep{ram} or a clean image of the scene background~\citep{loci_looped}. We instead propose to solve occlusion robustness and object permanence by:
(i) Object evidence modeling and ii) Position-appearance separation.

\begin{figure}[t]
\centering
\noindent\makebox[\linewidth][l]{\includegraphics{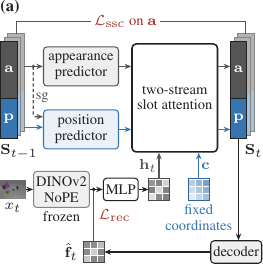}\hfill
\raisebox{0.02in}{\color[HTML]{B0B0B0}\rule{0.5pt}{1.72in}}\hfill
\includegraphics{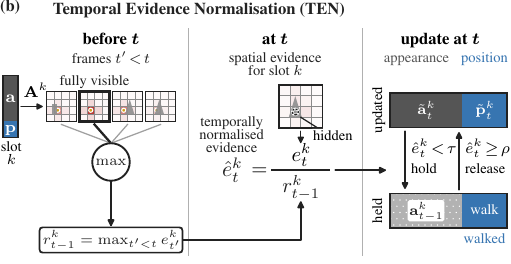}}
\caption{\textbf{(a) The two-stream slot in training. (b) \ten{} measures a slot's evidence on frame $t$
against the largest evidence the slot had before $t$.} (a) An appearance predictor and a position
predictor advance the two streams $\mathbf{a}$ and $\mathbf{p}$ of every slot; the position predictor
reads $\mathbf{a}$ through a stop-gradient (sg). The two-stream slot attention updates $\mathbf{a}$ from
the appearance grid $\mathbf{h}_t$ and $\mathbf{p}$ from the position grid $\mathbf{c}$
(Eq.~\ref{eq:splitkey}). The decoder places each object from $\mathbf{p}$ alone, and
$\mathcal{L}_{\text{ssc}}$ acts on $\mathbf{a}$ only (Sec.~\ref{sec:method:arch}). (b) For slot $k$,
$e^k_t$ is the mean of its largest attention values on frame $t$ (Eq.~\ref{eq:evidence}), and
$\hat e^k_t = e^k_t / r^k_{t-1}$ (Eq.~\ref{eq:tnorm}). The slot switches to \emph{held}, its appearance
kept and its position walked (Sec.~\ref{sec:method:carry}), when $\hat e^k_t$ falls below $\tau$, and back
to \emph{updated} when it rises above $\rho$ (Eq.~\ref{eq:hold}).}
\label{fig:norm}
\end{figure}

First, we add an explicit object evidence modeling that can acknowledge occlusions to act accordingly. An \emph{evidence} establishes the object's visibility, and can be obtained from a slot's attention to all patches in that frame. Although, previous methods compute the same attention, it is normalized spatially, thus deleting any occlusion cues. We instead obtain the correct evidence essentially free, by normalizing the attention with its maximum along a duration. This temporal evidence normalization (\ten{}) is applied only at inference. More importantly, \ten{} allows us to add an explicit \emph{hold} operation, as shown in Fig.~\ref{fig:norm}(b) to pause the slot update while the object is occluded (Fig. \ref{fig:teaser}(a), green line). We tackle the second sub-problem of position-appearance separation with a two stream (appearance and position) framework, hence the \emph{hold} operation in \ten{} happens solely on the appearance stream. In contrast, the hidden object's position stream continues its update with a trained \emph{walker}~\citep{ram}.
The two streams are shown in Fig.~\ref{fig:norm}(a), and consists of an appearance stream that reads DINOv2 features computed without a positional encoding \citep{dinosaw}, and a position stream that reads only the patch coordinates.
Finally in order to keep object permanence, we train the walker from \ten{}'s occlusion and reappearance detections (Fig.~\ref{fig:teaser}(a)bottom). We show the contributions of \ten{}, the two streams and the walker in permanence metrics ($8.6 \to 80.5$), with the three arrows of Fig.~\ref{fig:teaser}(b). The resulting model, \method{}, returns reappearing objects to their slot and tracks them while hidden, with no label used at any stage. Additionally, we show the potential of \method{} as a downstream planner of object slots.

Object permanence is considered as a key attribute of visual intelligence in cognitive science, with many studies in infants, in visual cortex and ideations in early computer vision research. Piaget~\citep{Piaget1954construction} made the search for an occluded object the milestone of infancy. LA-CATER~\cite{cater,lacater} formulates a computer vision benchmark on the same task. In early computer vision research, Marr's sketches made occluding contours and amodal completion explicit \citep{Marr1982vision,Nitzberg19902_1D} with similar reasoning. More relevant to our self-supervised training, Gibson~\citep{Gibson1979ecological} argued that human perception distinguishes occlusions from going out of existence and that observers learn the difference from reappearances. In this work, we model the requirements to learn object permanence self-supervised, \ie, the evidence modeling with \ten{}, the two-stream representation's explicit what/where separation, and the hidden location walker trained on reappearance in Gibson's principle and showcase its success in LA-CATER.

We summarize our contributions as:
\begin{itemize}
\item \textbf{Object evidence modelling.} Temporal evidence normalisation (\ten{}), which compares a slot's
attention with its own past to detect when its object is occluded, and holds the slot until the object reappears.
\item \textbf{Appearance-position separation.} A slot with separate appearance and position streams, so that the
appearance of an occluded object is held while its position keeps changing.
\item \textbf{Object permanence from reappearance.} A walker that follows the position of an occluded object,
trained only on where the object reappears, with no labels at any stage.
\end{itemize}

\section{Related Work}

\paragraph{Unsupervised object-centric video representation learning.}
Object-centric models learn, mostly without labels, to represent a video as a set of slots, each
bound to image features by slot attention \citep{slotattention} and initialised from the slots of the
previous frame \citep{savi,steve}. Trained to reconstruct pixels, flow or image tokens, they were shown
mostly on synthetic scenes, until self-supervised DINO features \citep{dino,dinov2} scaled them to
real images \citep{dinosaur} and real video \citep{videosaur,slotcontrast}. Later methods revise the
objective, the slot count or the queries \citep{srl,ssync,slotcurri,gc,smoothsa,randsfq}, and dynamics
models predict how the slots evolve \citep{slotformer,ocvp,statm,slotssm,ock}. Occlusion, however,
remains open: VideoSAUR reports slots reassigned to other objects in longer videos with occlusions.

\paragraph{Separating appearance from position.}
Generative scene models keep where an object is apart from what it looks like, with an explicit
position per object \citep{air,spair,space,scalor,gswm,loci} or per particle \citep{dlp,ddlp,lpwm},
which DDLP~\citep{ddlp} tracks over time and LPWM~\citep{lpwm} moves with a latent action per particle. A particle, however,
need not follow one object: DDLP reports particles shared by two objects and large objects split over
several. LPWM ties each particle to its patch of origin, and all three name camera motion as a
limitation. On DINO features, STAITUS \citep{staitus} splits the slot into appearance, position and
scale on ISA's per-slot reference frames \citep{isa}, reconstructing RGB from slots placed by their
pose. DINO features, however, carry position \citep{dinosaw}, so its appearance part is not free of it.

\paragraph{Object permanence.}
Object permanence has mostly been learnt from labels. OPNet on LA-CATER \citep{lacater}, PermaTrack
\citep{tokmakov2021permanence} and TCOW \citep{tcow} are trained on the positions of hidden objects,
and RAM \citep{ram} reduces the labels to boxes and track identities on visible frames, localising the
hidden object with a random walk along its memory that is supervised where the object reappears, on the
identity it had before the occlusion. Within object-centric
learning, Loci-Looped \citep{loci_looped} and OC-MOT \citep{ocmot} handle occlusion with little
supervision: a clean background image and detection labels on at most 6.25\,\% of frames, respectively.

\section{Method}
\label{sec:method}

A video $x_{1:T}$ shows a scene of at most $K$ objects. Each object has an
appearance, which changes slowly over the clip, and a position, its location and extent in the image,
which can change on every frame. An object can be fully occluded over a contiguous run of frames,
behind an occluder, inside a container or carried by another object, and then reappear
\citep{lacater}.

\paragraph{Object-centric video model.} A frozen self-supervised backbone $f$ \citep{dinov2} and a
small MLP $g_\psi$ map frame $x_t$ to $N$ patch features
$\mathbf{h}_t=g_\psi(f(x_t))\in\mathbb{R}^{N\times D}$. $K$ slots
$\mathbf{S}_t=\{\mathbf{s}^k_t\}_{k=1}^K$ represent the scene: a predictor $P_\omega$ advances them to
the next frame, and a corrector $C_\theta$ binds them to its features with slot attention
\citep{slotattention,savi},
\begin{equation}
\begin{split}
&\mathbf{S}^p_t=P_\omega(\mathbf{S}^c_{t-1}),\qquad
\mathbf{S}^c_t=C_\theta(\mathbf{h}_t,\mathbf{S}^p_t):\\
&\qquad\mathbf{A}_t=\operatorname{softmax}_K\!\Big(\tfrac{1}{\sqrt{D}}\,k(\mathbf{h}_t)\,q(\mathbf{S})^\top\Big),\quad
\mathbf{s}^k\leftarrow\mathrm{GRU}\Big(\sum_n \bar A^k_{t,n}\,v(\mathbf{h}_{t,n}),\ \mathbf{s}^{k}\Big),
\end{split}
\label{eq:sa}
\end{equation}
where the softmax is over slots, $\bar{\mathbf{A}}$ is $\mathbf{A}$ renormalised over patches, and the
update is iterated a few times per frame, starting from $\mathbf{S}=\mathbf{S}^p_t$; $\mathbf{S}^c_t$ is
its result. A decoder reconstructs the backbone features,
$\hat{\mathbf{f}}_t=\mathrm{Dec}(\mathbf{S}^c_t)$, with per-slot alpha masks that serve as the
segmentation \citep{dinosaur}. Training minimises feature reconstruction loss $\mathcal{L}_{rec}$ and a slot--slot contrastive
term $\mathcal{L}_{ssc}$ \citep{slotcontrast}, which makes a slot most similar to itself on the previous frame among all
slots in the batch,
\begin{equation}
\mathcal{L}=\sum_t \mathcal{L}_{\text{rec}}(f(x_t),\hat{\mathbf{f}}_t)
+\alpha\,\mathcal{L}_{\text{ssc}}(\mathbf{S}^c_{t-1},\mathbf{S}^c_t),
\qquad
\mathcal{L}_{\text{rec}}=\|f(x_t)-\hat{\mathbf{f}}_t\|_2^2 .
\label{eq:objective}
\end{equation}

Two properties of Eq.~\ref{eq:sa} lose an object under occlusion: i) The slot is one vector, so an update
reaches the object's appearance and its position together. ii) The renormalisation over patches gives
every slot an update of full weight, so the slot of a hidden object is rewritten with the patches at
its former location. \method{} splits the slot into two streams (Sec.~\ref{sec:method:arch}), then
makes the normalisation aware of temporal evidence, a primary prerequisite for handling occlusions,
and moves the position of a held slot with a walker (Sec.~\ref{sec:method:latch}).

\subsection{Two streams}
\label{sec:method:arch}

A slot is split into an appearance stream and a position stream (Fig.~\ref{fig:norm}a),
$\mathbf{s}^k=[\mathbf{a}^k;\mathbf{p}^k]$ with $\mathbf{p}^k\in\mathbb{R}^{4}$. Only $\mathbf{p}^k$ is
written from the position grid, and the decoder places an object from $\mathbf{p}^k$ alone.

\paragraph{Inputs.} The corrector reads two grids over the same $N$ patches (Fig.~\ref{fig:norm}a,
bottom). The appearance grid
$\mathbf{h}_t$ is computed from foundation-model features without a positional encoding
(Sec.~\ref{sec:exp:setup}). The position grid
$\mathbf{c}\in\mathbb{R}^{N\times 4}$ is a fixed Fourier code of each patch's image coordinate
$\mathbf{x}_n$ \citep{fourierfeatures}, the same on every frame.

\paragraph{Two-stream slot attention.} The attention logit of patch $n$ for slot $k$ is the sum of an
appearance term and a position term. There is one softmax over slots and one attention map, and each
stream is updated from its own grid by its own GRU:
\begin{equation}
\begin{split}
M^k_{t,n}&=q_a(\mathbf{a}^k)^{\!\top}k_a(\mathbf{h}_{t,n})+q_p(\mathbf{p}^k)^{\!\top}k_p(\mathbf{c}_n),\\
\mathbf{a}^k&\leftarrow\mathrm{GRU}_a\Big(\sum_n\bar A^k_{t,n}\,v_a(\mathbf{h}_{t,n}),\mathbf{a}^k\Big),\quad
\mathbf{p}^k\leftarrow\mathrm{GRU}_p\Big(\sum_n\bar A^k_{t,n}\,v_p(\mathbf{c}_n),\mathbf{p}^k\Big),
\end{split}
\label{eq:splitkey}
\end{equation}
with $\mathbf{A}_t=\operatorname{softmax}_K(M_t)$. A slot claims a patch whose content matches its
appearance and whose location matches its position. Because $\mathbf{p}^k$ is written from an
attention-weighted mean of fixed codes, it is the slot's location and extent on the grid, and the
recurrence carries it as a state. ISA~\citep{isa} instead recomputes position and
scale from the attention at every iteration.

\paragraph{Predictors and objective.} Appearance and position streams have their own predictors, each a
transformer over that stream of all slots. The position predictor is also conditioned on an 8-dimensional
projection of $\mathbf{a}$ under a stop-gradient (sg in Fig.~\ref{fig:norm}a), since an object's next
position depends on its shape. $\mathcal{L}_{\text{ssc}}$ acts on $\mathbf{a}$ only; the objective is
otherwise unchanged. Nothing in the architecture mixes the two streams' slot vectors.
Appendix~\ref{app:twostreams} details the decoder.

\subsection{Temporal evidence normalisation and the walker}
\label{sec:method:latch}

An occlusion changes whether an object is seen, not the object. Over an occlusion on frames
${t_s\le t<t_e}$, (i) the object's appearance persists and (ii) its position follows a continuous path,
so a reappearance is the same appearance seen at the end of that path. \ten{} detects $t_s$ and $t_e$
from the slot's own attention and holds its appearance, for (i) and the walker continues its position
along the path, for (ii).

\paragraph{Evidence modelling.} Detecting $t_s$ and $t_e$ needs two readings of the slot's attention:
whether the object is still where the slot last saw it, to start the hold, and whether it is seen
again anywhere, even partly, to release it. The first reading must be local: when its object is
hidden, a slot can move onto the occluder or another object, and its attention over the whole frame
stays high. How much attention a slot receives also differs between slots, objects and scenes, so a
threshold that fits one slot does not fit another; each reading is therefore compared with the same
slot's earlier values. Let $A^k_{t,n}$ be the share of patch $n$ that slot $k$ receives on frame $t$
under the softmax over slots (Eq.~\ref{eq:sa}), on the first corrector iteration, before the frame
changes the slots. The \emph{evidence}, read by the release, and the \emph{local evidence}, read by
the hold, are
\begin{equation}
e^k_t=\frac{1}{m}\sum_{n\in\mathcal{I}^k_t}A^k_{t,n},\qquad
e^{k,\mathrm{loc}}_t=\max_n\frac{1}{|\mathcal{B}(n)|}\sum_{n'\in\mathcal{B}(n)}W^k_{n'}\,A^k_{t,n'},
\label{eq:evidence}
\end{equation}
with $\mathcal{I}^k_t$ the $m$ patches of largest $A^k_{t,n}$; $\mathcal{B}(n)$ the $3\times3$ block
around patch $n$, which asks for a connected piece of the object; and $W^k$ a window built from the
slot's final-iteration attention on its last frame not held (Appendix~\ref{app:method}).
Each is divided by its own largest value on the earlier frames of the clip (Fig.~\ref{fig:norm}b,
left),
\begin{equation}
\hat e^k_t=\frac{e^k_t}{r^k_{t-1}},\ \ r^k_{t-1}=\max_{t'<t}e^k_{t'};\qquad
\hat e^{k,\mathrm{loc}}_t=\frac{e^{k,\mathrm{loc}}_t}{r^{k,\mathrm{loc}}_{t-1}},\ \
r^{k,\mathrm{loc}}_{t-1}=\max_{t'<t}e^{k,\mathrm{loc}}_{t'}.
\label{eq:tnorm}
\end{equation}
A slot is \emph{held} from the frame on which $\hat e^{k,\mathrm{loc}}_t$ falls below $\tau$ until the
frame on which $\hat e^k_t$ rises above $\rho>\tau$, so a partial glimpse does not end it
(Fig.~\ref{fig:norm}b, right); a hold does not start
while $e^{k,\mathrm{loc}}_t$ is rising sharply (Appendix~\ref{app:method}). With
$u^k_t\in\{0,1\}$ the held indicator, $\tilde{\mathbf{a}}^k_t$ and $\tilde{\mathbf{p}}^k_t$ the
corrector's output and $\hat{\mathbf{p}}^k_t$ the walker's position,
\begin{equation}
\begin{split}
u^k_t&=\mathbf{1}\big[\hat e^{k,\mathrm{loc}}_t<\tau\big]\,(1-u^k_{t-1})+\mathbf{1}\big[\hat e^k_t<\rho\big]\,u^k_{t-1},\\
\mathbf{a}^k_t&=(1-u^k_t)\,\tilde{\mathbf{a}}^k_t+u^k_t\,\mathbf{a}^k_{t-1},\qquad
\mathbf{p}^k_t=(1-u^k_t)\,\tilde{\mathbf{p}}^k_t+u^k_t\,\hat{\mathbf{p}}^k_t.
\end{split}
\label{eq:hold}
\end{equation}

\paragraph{The walker.}
\label{sec:method:carry}
The walker follows the content at a held slot's last position and shifts that position by how far the
content moves. It follows the content with a random walk over patches on a graph of appearance
features \citep{crw,ram}. To read a patch in the coordinates of the position stream, let
$P^*(n)$ be the position of a slot that attends to patch $n$ alone: the value to which $\mathrm{GRU}_p$
(Eq.~\ref{eq:splitkey}) converges when its input is $v_p(\mathbf{c}_n)$, with $\mathbf{c}_n$ the
Fourier code of the patch's coordinate (Sec.~\ref{sec:method:arch}). Let $n_a$ be the patch whose
$P^*$ is nearest the slot's position $\mathbf{p}^k_{t_0}$ on an updated frame $t_0<t_s$ (Appendix~\ref{app:method}). The walk $\pi_t$, a distribution over patches, starts as a point mass at
$n_a$ and moves between consecutive appearance grids,
\begin{equation}
\pi_t=\pi_{t-1}\,\mathcal{T}_t,\qquad
\mathcal{T}_t(n\!\to\!n')=\operatorname{softmax}_{n'\in\mathcal{W}(n)}\!\big(s\,\langle\mathbf{z}_{t-1,n},\mathbf{z}_{t,n'}\rangle+\beta_{n'-n}\big),
\label{eq:transition}
\end{equation}
with $\mathcal{W}(n)$ a window around $n$, $\mathbf{z}_t$ a learned embedding of $\mathbf{h}_t$, $s$ a
learned scale and $\beta$ a learned prior over offsets. The walker's position is
$\hat{\mathbf{p}}^k_t=\mathbf{p}^k_{t_0}+P^*(\bar{\mathbf{x}}_t)-P^*(n_a)$, with $P^*$ interpolated at
$\bar{\mathbf{x}}_t$, the mean position of $\pi_t$ near its mode. As the held slot's position
(Eq.~\ref{eq:hold}), $\hat{\mathbf{p}}^k_t$ places the slot's next attention, so the kept appearance is
looked for where the walker expects the object. The walker is trained on reappearance: its only target
is $\mathbf{p}^k_{t_e+1}$, the position the encoder gives the slot just after its release at $t_e$,
\begin{equation}
\mathcal{L}_{\text{walk}}=-(1+\lambda L)\,\log\sum_n\pi_{t_e-1}(n)\;
\mathcal{N}\!\big(\mathbf{p}^k_{t_e+1};\ \mathbf{p}^k_{t_0}+P^*(n)-P^*(n_a),\ \sigma^2\mathbf{I}\big),
\label{eq:carryloss}
\end{equation}
with $\mathcal{N}$ a Gaussian of learned width $\sigma$, $L$ the length of the occlusion and $\lambda$ a
weight per occluded frame. Because the held appearance returns the object to the slot that held it,
the target is the right object's, and no ground-truth position enters the loss.

The encoder is trained once with Eq.~\ref{eq:objective} and \ten{} has no parameters; the walker is
trained afterwards on the encoder's outputs, and at inference \ten{} and the walker run together. Appendix~\ref{app:method} gives the stages
(Table~\ref{tab:stages}), our method's inference algorithm (Algorithm~\ref{alg:inference}) and other constants.

\section{Experiments}

\subsection{Setup}
\label{sec:exp:setup}

\paragraph{Implementation details.}
The backbone is DINOv2 ViT-S/14 with registers \citep{dinov2,registers} without its positional
embedding \citep{dinosaw}, frozen. Slots have width 64, split into 60 appearance and 4 position
dimensions; $K=11$ on LA-CATER and 15 on KITTI. Everything else is the configuration of
\citet{slotcontrast}, except the training window, which is 8 frames rather than 4 so that an occlusion
and the reappearance after it fall inside one window. The dynamics model is OCVP-Par \citep{ocvp}. Every other value is in Appendices~\ref{app:method}, \ref{app:setup} and~\ref{app:dynmodel}.

\paragraph{Datasets.}
LA-CATER \citep{lacater} extends CATER \citep{cater}: rendered scenes in which geometric objects move
and occlude one another, and a target object is covered by cones that may then carry it. Every object
is labelled on every frame as visible, occluded, contained or carried, with an amodal box while it is
occluded. We use its own static camera and the moving camera of \citet{ram}, and score 195 and 388
occlusions of the target, 183 and 360 of them with a scorable reappearance (Appendix~\ref{app:datasets}). MOVi-C \citep{kubric} renders scanned objects falling on photographic
backgrounds and carries discovery, separation and dynamics on its 250-video validation split, as in
\citet{slotcontrast}. KITTI \citep{kitti} is real-world driving video, split in time into training and
validation halves as in \citet{ram}; discovery is scored on the 70 validation clips and
re-identification on the occlusions of cars, both from the masks of KITTI-MOTS \citep{mots}.

\paragraph{Baselines.}
We build on top of SlotContrast \citep{slotcontrast}. We train it under same settings except the training clip window, which is changed from 4 to 8 frames to capture more occlusion-reappearance events. RandSF.Q \citep{randsfq}
predicts the next frame's slot-attention queries from slot-feature pairs sampled from the recurrence.
We run the authors' code with a LA-CATER configuration of our own. Loci-Looped \citep{loci_looped},
the permanence model closest to ours, keeps a position and a gestalt code per slot and imagines the
position while its percept gate is closed. It also requires a background image of the scene, which its
released loader builds from the ground-truth masks, so its rows are not label-free. RAM \citep{ram} is
the supervised reference, trained with boxes, track identities and visibility labels. We report its raw output
(Appendix~\ref{app:baselines}). \emph{Frozen} is \method{}'s slot frozen at its last visible
position.

\paragraph{Metrics.}
Permanence is the mAP of OPNet \citep{lacater}: the share of an occlusion's hidden frames on which the
predicted box overlaps the target's amodal box above an IoU threshold, averaged over occlusions. We
report \emph{mAP@$[0.1,0.3]$} $\uparrow$, its mean over the thresholds 0.10 to 0.30. A slot model's box
keeps the size of its mask before the occlusion and is centred on the predicted position, as in
\citet{ram} (Appendix~\ref{app:permmetric}).

Re-identification is \emph{same-slot return} (SAME) $\uparrow$: the share of occlusions on which the
slot that owns the target at the reappearance is the slot that owned it before the occlusion. A slot
owns the target on a frame when its grouping attention carries the largest mass on the target's visible
pixels. Before and after are the last and the first frame on which at least half of the target's
amodal mask is visible. Appendix~\ref{app:hota} reports HOTA \citep{hota} over whole clips.

Discovery uses video FG-ARI \citep{ari} and video mBO on the decoder's masks over the whole clip, as
in \citet{slotcontrast}.

\subsection{Object permanence and re-identification}

\begin{table}[t]
\begin{minipage}[t]{0.575\linewidth}
\vspace{0pt}
\centering\scriptsize
\setlength{\tabcolsep}{2.1pt}%
\caption{\textbf{Position of the object under occlusion in the LA-CATER dataset.} mAP@$[0.1,0.3]$
$\uparrow$ ($\times100$) on the hidden frames of each label. In parentheses, what a row is given
beyond the video (bg: a per-frame background image); \emph{frozen} holds \method{}'s slot at its last
visible position. Under the static camera \method{} is close to the supervised method RAM. Bold:
best row that reads the video alone, which excludes \emph{frozen}, RAM and Loci-Looped.}
\label{tab:permanence}
\begin{tabular}{@{}l ccc ccc@{}}
\toprule
& \multicolumn{3}{c}{static camera} & \multicolumn{3}{c}{moving camera} \\
\cmidrule(lr){2-4}\cmidrule(lr){5-7}
& occluded & contained & carried & occluded & contained & carried \\
\midrule
\privrow \priv{RAM (vis., bbox, id)} & \priv{73.8} & \priv{86.7} & \priv{81.0} & \priv{83.2} & \priv{87.4} & \priv{81.1} \\
Loci-Looped (bg)                 & 56.6 & 49.9 & 32.9 & 49.5 & 51.8 & 37.9 \\
frozen                           & 57.8 & 56.8 & 41.9 & 51.4 & 32.2 & 13.6 \\
\midrule
SlotContrast                     & 8.6  & 4.3  & 3.5  & 3.1  & 1.3  & 1.2 \\
RandSF.Q                         & 11.5 & 6.4  & 5.4  & 1.3  & 1.2  & 1.0 \\
\method{}                        & \textbf{80.5} & \textbf{84.2} & \textbf{78.7} & \textbf{54.4} & \textbf{51.6} & \textbf{46.3} \\
\bottomrule
\end{tabular}
\end{minipage}\hfill
\begin{minipage}[t]{0.395\linewidth}
\vspace{0pt}
\centering
\includegraphics[width=\linewidth]{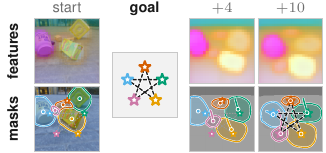}
\captionof{figure}{\textbf{Planning in imagination on MOVi-C: the goal is a star of the clip's own objects.}
\emph{goal}: a pentagon. \emph{features} (top): decoded DINO features,
PCA as RGB; \emph{masks} (bottom): decoded slot masks with their tracks. Action optimisation by
gradient descent: the error falls from $26.7$ to $1.5$\,px.}
\label{fig:dynamics}

\end{minipage}
\end{table}

\begin{figure}[t]
\centering
\includegraphics[width=\textwidth]{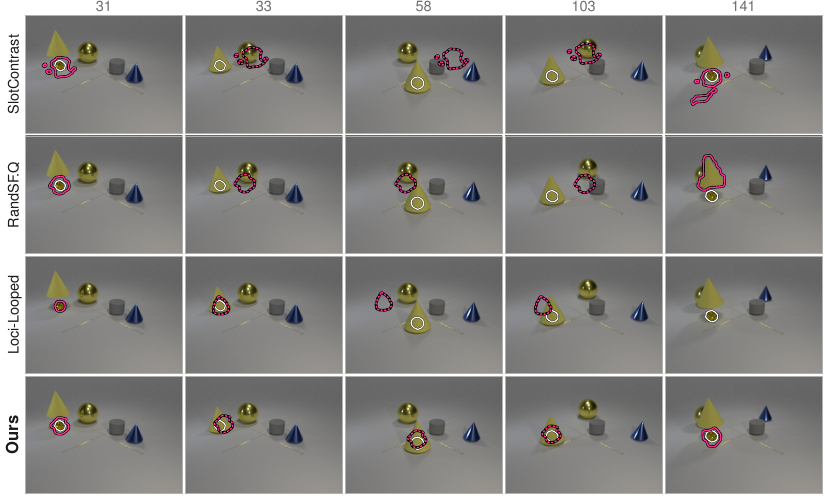}
\caption{\textbf{Object permanence in LA-CATER: a cone is put over the target and carries it across the frame.}
The target is visible only in the first and last columns. White: its amodal mask. Red: the followed
slot, \emph{solid} for its own mask, \emph{dashed} where moved onto the predicted position; no red: the slot
owns no pixel there. Only \method{} tracks the target while it is hidden.}
\label{fig:qual-perm}
\end{figure}

In Table~\ref{tab:permanence}, we show that \method{}, with no label at any stage, is within 4.1 points
of the supervised RAM under the static camera (80.9 vs 85.0 over all hidden frames) and ahead of every
other model. Carried frames are the
hardest case, because the container moves the target away from its last visible position:
\emph{frozen} falls from 57.8 on occluded frames to 41.9 on carried ones, and Loci-Looped from 56.6
to 32.9, while \method{} keeps 78.7. Fig.~\ref{fig:qual-perm} shows one such occlusion, and more in
Appendix~\ref{app:qual}. Under the moving camera, \method{} scores lower than in static camera. The drop starts upstream of the walker, in the encoder, which discovers objects less well
under this camera and returns fewer targets to their slot.

\begin{table}[t]
\centering\scriptsize
\setlength{\tabcolsep}{3.1pt}
\caption{\textbf{Re-identification after an occlusion, LA-CATER \& KITTI.} SAME $\uparrow$ (\%) by the length
of the occlusion in seconds. \emph{requires}: what the row is given
beyond the video, at training or at test time. With no label at any stage \method{} is ahead of every row
but the supervised RAM in every bin, and ahead of RAM as well on the 0.2--1.0\,s and 2.1--4.2\,s occlusions
under the static camera.}
\label{tab:same}
\resizebox{\linewidth}{!}{%
\begin{tabular}{@{}l l cccc cccc c@{}}
\toprule
& & \multicolumn{4}{c}{LA-CATER, static camera} & \multicolumn{4}{c}{LA-CATER, moving camera} & KITTI \\
\cmidrule(lr){3-6}\cmidrule(lr){7-10}\cmidrule(lr){11-11}
& requires & 0.2--1.0\,s & 1.0--2.1\,s & 2.1--4.2\,s & $\geq$4.2\,s & 0.2--1.0\,s & 1.0--2.1\,s & 2.1--4.2\,s & $\geq$4.2\,s & 0.1--2.2\,s \\
\midrule
\privrow \priv{RAM}      & \priv{boxes, identities, visibility} & \priv{90.2} & \priv{95.5} & \priv{85.7} & \priv{88.7} & \priv{95.0} & \priv{90.0} & \priv{77.3} & \priv{69.5} & \priv{87.1} \\
Loci-Looped              & per-frame background & 53.1 & 33.3 & 24.5 & 25.4 & 39.5 & 31.1 & 27.3 & 21.1 & -- \\
\midrule
RandSF.Q                 & none & 56.1 & 22.7 & 16.3 & 21.1 & 38.1 & 23.3 & 16.7 & 15.8 & -- \\
SlotContrast             & none & 63.4 & 27.3 & 20.4 & 23.9 & 30.2 & 11.7 & 12.1 & 13.7 & 29.0 \\
\method{}                & none & \textbf{95.1} & \textbf{86.4} & \textbf{91.8} & \textbf{77.5} & \textbf{82.0} & \textbf{81.7} & \textbf{75.8} & \textbf{64.2} & \textbf{48.4} \\
\bottomrule
\end{tabular}}
\end{table}

\begin{figure}[t]
\centering
\includegraphics[width=\textwidth]{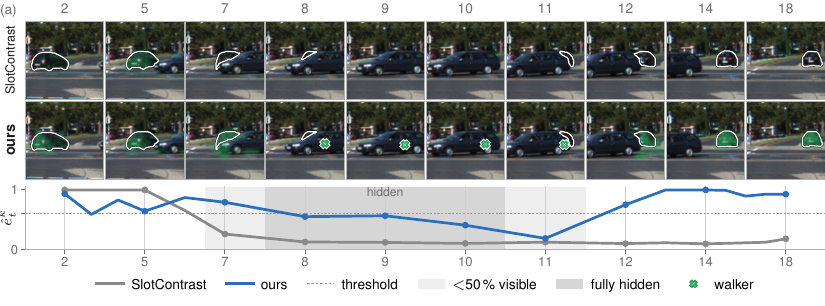}
\caption{\textbf{Occlusions in KITTI: a car passes behind another road user and reappears.}
Unevenly spaced frames of one tile, cropped to the car and its occluder. White: the car's visible mask.
Green: the slot on the car before the occlusion; cross: the walker's predicted centre while the car is hidden.
Both models' evidence falls while the car is hidden; only \method{}'s recovers, and its slot returns.}
\label{fig:kitti-reid}
\end{figure}

Table~\ref{tab:same} gives SAME by the length of the occlusion, and Fig.~\ref{fig:kitti-reid} shows
one occlusion on KITTI. Figs.~\ref{fig:qual-reid} and~\ref{fig:kitti-reid-supp} give
more examples. Under the static camera, every baseline except the supervised RAM loses the target's
slot once the occlusion exceeds a second: its SAME falls by 20 to 36 points and does not recover.
\method{} still returns the target after 77.5\,\% of occlusions longer than four seconds under the
static camera and 64.2\,\% under the moving one, where these baselines reach at most 25.4\,\%. On
KITTI, it returns the car to its slot after 48.4\,\% of occlusions, against 29.0\,\% for
SlotContrast.

\subsection{Object discovery}

\begin{table}[t]
\centering\scriptsize
\setlength{\tabcolsep}{4pt}
\caption{\textbf{Object discovery on MOVi-C, LA-CATER and KITTI.} Video FG-ARI $\uparrow$ / video mBO
$\uparrow$ ($\times100$) on decoder masks over whole clips. Splitting the slot costs no discovery.}
\label{tab:discovery}
\resizebox{\linewidth}{!}{%
\begin{tabular}{@{}l l cc cc cc cc@{}}
\toprule
& & \multicolumn{2}{c}{MOVi-C} & \multicolumn{2}{c}{LA-CATER static} & \multicolumn{2}{c}{LA-CATER moving} & \multicolumn{2}{c}{KITTI} \\
\cmidrule(lr){3-4}\cmidrule(lr){5-6}\cmidrule(lr){7-8}\cmidrule(lr){9-10}
method & features & FG-ARI $\uparrow$ & mBO $\uparrow$ & FG-ARI $\uparrow$ & mBO $\uparrow$ & FG-ARI $\uparrow$ & mBO $\uparrow$ & FG-ARI $\uparrow$ & mBO $\uparrow$ \\
\midrule
SlotContrast             & DINOv2      & 66.9 & 32.0 & 94.2 & 21.5 & 83.1 & 13.7 & 42.4 & 12.4 \\
SlotContrast + \ten{}    & DINOv2      & 68.1 & 31.4 & 94.5 & 22.3 & 86.7 & 14.7 & 39.4 & 12.3 \\
\method{}                & DINOv2 NoPE & \textbf{71.6} & \textbf{36.0} & \textbf{96.6} & \textbf{23.9} & \textbf{89.4} & \textbf{17.8} & \textbf{48.9} & \textbf{22.9} \\
\bottomrule
\end{tabular}}
\end{table}

We report object discovery in Table~\ref{tab:discovery}, and show the masks and slot identities on
MOVi-C and KITTI in Figs.~\ref{fig:qual-disc-movic} and~\ref{fig:qual-disc-kitti}.
\method{} discovers objects better than SlotContrast on every dataset.
Separating appearance from position matters more on real video: with \ten{} on both encoders,
\method{} leads by 9.5 FG-ARI and 10.5 mBO on KITTI, against 2.1 to 3.5 FG-ARI on the rendered
datasets, while \ten{} alone does not help SlotContrast there (42.4 to 39.4 FG-ARI).

\subsection{Application: planning with a world model}
\label{sec:exp:dynamics}

A world model predicts how the world changes under an action. It must learn to
separate the action from everything else that changes. Without the action as input, the future is ambiguous, and a
predictor trained under a squared error averages over the possible futures \citep{mathieu2016}.
SlotFormer \citep{slotformer} and OCVP \citep{ocvp} do exactly that \citep{playslot}.

In \method{}, the action comes for free. The position stream is the object's kinematic state, and its
change is the action. Given to the predictor through a two-dimensional bottleneck (Fig.~\ref{fig:dyn-arch}), it becomes a code of the object's motion. SlotContrast's whole-slot change,
given the same way, does not, and its predictor stays 9.4 FG-ARI below its oracle, against 3.9 for
\method{} (Table~\ref{tab:dynamics}).

Because the rollout is differentiable in the action, planning follows by gradient descent. We set the goal to
arrange a clip's objects into a star, the planner brings their mask centres from 25.1 to 5.2\,px of
their vertices at 128\,px over 20 validation clips (Fig.~\ref{fig:dynamics}).

\subsection{Ablations}

\begin{table}[t]
\begin{minipage}[t]{0.48\linewidth}
\vspace{0pt}
\centering\scriptsize
\setlength{\tabcolsep}{3pt}
\caption{\textbf{Ablation on LA-CATER (static).} Each row adds one component to the row above,
starting from SlotContrast. mAP is the mAP@$[0.1,0.3]$ of Table~\ref{tab:permanence}; \emph{all}: every
frame of the clip, the slot fixed before the occlusion. Bold: best per column.}
\label{tab:ablation}
\begin{tabular}{@{}l c cc cc@{}}
\toprule
& & \multicolumn{2}{c}{discovery} & \multicolumn{2}{c}{mAP $\uparrow$} \\
\cmidrule(lr){3-4}\cmidrule(lr){5-6}
method & SAME $\uparrow$ & FG-ARI $\uparrow$ & mBO $\uparrow$ & all & hidden \\
\midrule
SlotContrast              & 32.2 & 94.2 & 21.5 & 14.6 & 6.6 \\
\quad + \ten{}            & 77.6 & 94.5 & 22.3 & 22.9 & 19.6 \\
\quad + two streams       & 71.0 & 95.3 & 18.5 & 20.3 & 6.9 \\
\quad + NoPE features     & 82.0 & 96.6 & 23.9 & 69.0 & 30.1 \\
\quad + walker (\method{}) & \textbf{86.3} & \textbf{96.9} & \textbf{24.0} & \textbf{88.2} & \textbf{80.9} \\
\bottomrule
\end{tabular}
\end{minipage}\hfill
\begin{minipage}[t]{0.50\linewidth}
\vspace{0pt}
\centering\scriptsize
\setlength{\tabcolsep}{5pt}
\caption{\textbf{Dynamics predictor on MOVi-C with action input.} Video FG-ARI $\uparrow$ / video mBO
$\uparrow$ ($\times100$) over ten rolled-out frames after a 14-frame burn-in; architecture in Fig.~\ref{fig:dyn-arch}. \emph{frozen} repeats the burn-in's last slots; \emph{oracle} encodes the true
future. \method{} comes close to its oracle; SlotContrast falls further short of its own.}
\label{tab:dynamics}
\renewcommand{\arraystretch}{1.14}
\begin{tabular}{@{}l cc@{}}
\toprule
model & FG-ARI $\uparrow$ & mBO $\uparrow$ \\
\midrule
\privrow \priv{oracle / frozen, SlotContrast} & \priv{76.0 / 55.0} & \priv{31.6 / 21.8} \\
\privrow \priv{oracle / frozen, \method{}}    & \priv{78.4 / 54.2} & \priv{37.2 / 23.6} \\
\midrule
SlotContrast & 66.6 & 26.9 \\
\method{}    & \textbf{74.5} & \textbf{33.1} \\
\bottomrule
\end{tabular}
\end{minipage}
\end{table}

\ten{} improves re-identification already on SlotContrast (Table~\ref{tab:ablation}). Splitting
appearance and position hurts while the features carry position. On features computed without a positional encoding
(DINOv2 NoPE, \citealp{dinosaw}), the split improves re-identification and, most notably, mAP. The walker
then tracks the object while it is hidden.

\section{Conclusion}
\method{} identifies two properties that object permanence requires of a slot representation. Building
on these properties, \method{} achieves object permanence without
any labels: it returns objects to their slots after long occlusions and localises them while hidden. Under a static camera, it is close to a tracker trained with boxes, track identities and visibility labels. The
separation of appearance and position gives world models a natural action, through which a planner reaches its goal by gradient
descent. Taken together, we expect these properties to carry over to other object-centric models, and
to make their representations a basis for world models that plan through occlusion.

\paragraph{Limitations.}
\method{} is built on its slot model and inherits its binding. When an object is occluded, its slot can
latch onto another object that looks alike (Fig.~\ref{fig:kitti-fail}a) or be released
onto one before the object returns (Fig.~\ref{fig:lacater-fail}), and a slot that is not on the target
when the occlusion begins cannot follow it (Fig.~\ref{fig:perm-fail}, \emph{wrong slot}). It also
inherits a fixed number of slots per dataset. Conversely, any advance in the underlying slot model
carries into \method{} for free, and with it into re-identification and permanence.

\subsubsection*{AI Use Statement}
We used AI assistants (Claude, Gemini, ChatGPT) to write the code, to search and cross-check related
work, and to edit the text. The ideas, the design of the experiments and the claims are the authors',
who checked them against the code and the cited papers.

\subsubsection*{Reproducibility Statement}
The encoder changes only the keys of the released SlotContrast configuration named in
Secs.~\ref{sec:method} and~\ref{sec:exp:setup}. Every other constant, the scored occlusions, the metrics,
the baselines' checkpoints and the dynamics model are in Appendices~\ref{app:method},
\ref{app:setup} and~\ref{app:dynmodel}. Code, configurations, the occlusion lists and the trained
walkers will be released.

\bibliography{references}
\bibliographystyle{iclr2027_conference}

\clearpage
\appendix
\setcounter{table}{0}\setcounter{figure}{0}
\renewcommand{\thetable}{S\arabic{table}}\renewcommand{\theHtable}{S\arabic{table}}
\renewcommand{\thefigure}{S\arabic{figure}}\renewcommand{\theHfigure}{S\arabic{figure}}
\section{Method}
\label{app:method}

\begin{table}[h]
\centering\small
\setlength{\tabcolsep}{5pt}
\begin{tabular}{@{}l>{\raggedright\arraybackslash}p{3.4cm}>{\raggedright\arraybackslash}p{4.0cm}l@{}}
\toprule
stage & what is trained & signal & what is frozen \\
\midrule
1. encoder & $g_\psi$, corrector, predictors, decoder & Eq.~\ref{eq:objective} & backbone $f$ \\
2. \ten{} & nothing & none & everything \\
3. walker & the walker of Eq.~\ref{eq:transition} & Eq.~\ref{eq:carryloss}, on the positions and holds of stages 1--2 & encoder, $P^*$ \\
\bottomrule
\end{tabular}
\caption{The three stages of \method{}. At inference, \ten{} and the walker run together, and the
walker's position is written into the held slots. No box, mask, visibility or occlusion label is used
at any stage.}
\label{tab:stages}
\end{table}

\subsection{Two streams}
\label{app:twostreams}
Slots are 64-dimensional, $[0,60)$ the appearance stream and $[60,64)$ the position stream, and are
initialised by SlotContrast's learned initialisation over all 64 dimensions. The position grid is
$\mathbf{c}_n=[\sin\tfrac{\pi}{2}\mathbf{x}_n;\cos\tfrac{\pi}{2}\mathbf{x}_n]$ with
$\mathbf{x}_n\in[-1,1]^2$ the patch centre, carried in its own channels. The position read-in maps
start at the identity, so the first position logit is a spotlight on the slot's position and the
first position update is the circular mean of the coordinate codes it attends to. The position stream
is RMS-normalised, has no MLP, and enters the attention at gain 1. The corrector runs two iterations
per frame and three on the first. Each predictor is one transformer block with four heads. The
decoder reconstructs $f(x_t)$ in each object's own frame \citep{isa}, reading a position
$\boldsymbol\mu^k$ and a scale $\boldsymbol\sigma^k$ from $\mathbf{p}^k$ alone:
\begin{equation*}
\begin{split}
(\boldsymbol\mu^k,\boldsymbol\sigma^k)&=h(\mathbf{p}^k),\qquad
\mathbf{u}^k_n=\frac{\mathbf{x}_n-\boldsymbol\mu^k}{\delta\,\boldsymbol\sigma^k},\\
(\hat{\mathbf{f}}^k_{n},\alpha^k_n)&=\mathrm{MLP}\big([\mathbf{a}^k;\gamma(\mathbf{u}^k_n)]\big),\qquad
\hat{\mathbf{f}}_n=\sum_k \operatorname{softmax}_k(\alpha^k_n)\,\hat{\mathbf{f}}^k_{n},
\end{split}
\end{equation*}
with $h$ a head of 64 hidden units ($\tanh$ for $\boldsymbol\mu$, a clamped exponential for
$\boldsymbol\sigma\in[0.01,4]$), $\delta=5$ as in \citet{isa}, $\gamma$ an 8-band Fourier embedding,
and an MLP of three hidden layers of 1024. The painting MLP has no access to absolute position.

\subsection{Temporal evidence normalisation}
The thresholds are $\tau=0.5$ and $\rho=0.9$. The evidence is read from the first corrector iteration and is detached; $m=4$. The window $W^k$ is
the slot's final-iteration attention on the last frame on which it was not held, scaled to a maximum of
one and dilated by a $5\times5$ maximum filter; $W^k=1$ on the first frame, which is never held. The
references $r^k$ and $r^{k,\mathrm{loc}}$ are both updated on held frames as well as on updated ones. A hold does not start while the local evidence is rising: besides
$\hat e^{k,\mathrm{loc}}_t<\tau$, it requires $\big(e^{k,\mathrm{loc}}_t/(2\,\bar e^{k}_{t-1})\big)^8<\tau$,
with $\bar e^k$ the exponential average of $e^{k,\mathrm{loc}}$ at rate $0.5$. A held slot keeps
dimensions $[0,60)$ of the state it committed on the previous frame; the decoder reads the state the
corrector produced for the current frame. On a slot that is one vector, the hold keeps the whole vector.

\subsection{Walker}
\label{app:walker}
$P^*$ is computed from the frozen corrector, by 200 iterations of $\mathrm{GRU}_p$ from zero driven by
$v_p(\mathbf{c}_n)$, and is interpolated bilinearly between patches. The walker moves every second
frame, the rate of the encoder's training clips; on the frame in between, the position written into
the slot is read one step ahead, from the next frame's appearance grid. It restarts whenever the slot
has been updated on three consecutive steps, from the slot's position on the middle one, so a hold is
walked from a position the slot had while its object was clearly visible.
$\bar{\mathbf{x}}_t$ is the $\pi_t$-weighted mean position over the $3\times3$
patches around the mode of $\pi_t$ smoothed by a $3\times3$ box. The embedding $\mathbf{z}_t$ is a
layer norm and a $1\times1$ convolution of $\mathbf{h}_t$ to 32 channels, plus a zero-initialised
residual from a $3\times3$ convolutional GRU \citep{convgru} run forward over the clip, followed by a $1\times1$
convolution and $\ell_2$ normalisation. $\mathcal{W}(n)$ is $11\times11$ patches; $s$ starts at 20,
$\beta$ at 5 on the zero offset and 0 elsewhere, and $\sigma$ at 0.05 with a floor of 0.02. In
Eq.~\ref{eq:carryloss}, $\lambda=1$ per frame. Anchors and targets are kept only where the slot's
position lies within a distance of $P^*$ set per encoder (a point-like slot), 0.35 on the static
camera and 0.6 on the moving one. The walker writes its position into a held slot only where
$\mathbf{p}^k_{t_0}$ lies within that distance of some $P^*(n)$; elsewhere
$\hat{\mathbf{p}}^k_t=\tilde{\mathbf{p}}^k_t$. Each clip adds up to
eight pseudo-occlusions of 1 to 8 steps on stretches where the slot is updated throughout; stretches
over which it moves two patches or more are drawn five times as often. The walker has about
$6\times10^4$ parameters and is trained on 1,900 clips with Adam at learning rate $10^{-3}$, batch 16, for 2,000 steps
of a 4,000-step schedule.

\begin{algorithm}[h]
\caption{\method{} inference, one frame $t$ (per slot $k$; all slots in parallel)}
\label{alg:inference}
\begin{algorithmic}[1]
\Require $\mathbf{h}_t$, $\mathbf{c}$, previous state $\mathbf{S}^c_{t-1}$, held indicator $u^k_{t-1}$, references $r^k_{t-1}$, $r^{k,\mathrm{loc}}_{t-1}$, window $W^k$, walker state $\pi$
\State $\mathbf{S}^p_t\gets P_\omega(\mathbf{S}^c_{t-1})$;\quad $\tilde{\mathbf{S}}_t,\mathbf{A}_t\gets C_\theta(\mathbf{h}_t,\mathbf{c},\mathbf{S}^p_t)$, $\mathbf{A}_t$ of the first iteration \Comment{Eqs.~\ref{eq:sa},~\ref{eq:splitkey}}
\State $e^k_t\gets$ top-$m$ mean of $\mathbf{A}^k_t$;\quad $e^{k,\mathrm{loc}}_t\gets$ largest $3{\times}3$ mean of $\mathbf{A}^k_t\odot W^k$ \Comment{Eq.~\ref{eq:evidence}}
\State $\hat e^k_t\gets e^k_t/r^k_{t-1}$,\ $\hat e^{k,\mathrm{loc}}_t\gets e^{k,\mathrm{loc}}_t/r^{k,\mathrm{loc}}_{t-1}$;\quad update both references by their maximum \Comment{Eq.~\ref{eq:tnorm}}
\State $u^k_t\gets$ held indicator from $\hat e^{k,\mathrm{loc}}_t$, $\hat e^k_t$, $u^k_{t-1}$ \Comment{Eq.~\ref{eq:hold}}
\If{$u^k_t=1$} \Comment{held}
  \State $\mathbf{a}^k_t\gets\mathbf{a}^k_{t-1}$ \Comment{appearance kept}
  \State $\pi\gets\pi\,\mathcal{T}_t$;\quad $\hat{\mathbf{p}}^k_t\gets\mathbf{p}^k_{t_0}+P^*(\bar{\mathbf{x}}_t)-P^*(n_a)$ \Comment{position walked, Eq.~\ref{eq:transition}}
  \State $\mathbf{p}^k_t\gets\hat{\mathbf{p}}^k_t$ \Comment{the next frame's attention, and so the release, looks for $\mathbf{a}^k$ here}
\Else \Comment{updated}
  \State $[\mathbf{a}^k_t;\mathbf{p}^k_t]\gets\tilde{\mathbf{s}}^k_t$;\quad restart $\pi$ at $\mathbf{p}^k_t$
  \State $W^k\gets$ the slot's final-iteration attention, scaled to a maximum of one and dilated
\EndIf
\State decode $\hat{\mathbf{f}}_t$ and masks from $\tilde{\mathbf{S}}_t$
\State \Return $\mathbf{S}^c_t$, which $P_\omega$ reads at $t+1$
\end{algorithmic}
\end{algorithm}

\section{Setup}
\label{app:setup}

\subsection{Implementation details}

\paragraph{Encoder.}
The backbone is the checkpoint released by \citet{dinosaw}: DINOv2 ViT-S/14 with registers, its learned
positional embedding deleted and the network fine-tuned on COCO images to reproduce the original patch
features. It is frozen at the resolution of each dataset: 336 px on MOVi-C, giving
$24\times24=576$ patches, as in \citet{slotcontrast}; 518 px on LA-CATER, giving
$37\times37=1369$; and 378 px on KITTI, giving $27\times27=729$. A two-layer MLP maps its
384-dimensional features to 64, as in \citet{slotcontrast}.

\paragraph{KITTI.}
Each frame is brought to $384\times1280$ by the affine map of \citet{ram} and resized uniformly to
$378\times1260$. A training window is 8 consecutive frames at 10\,fps, the recording rate, cut to
one $378\times378$ crop starting at $x\in\{0,294,588,882\}$, the crops of \citet{mrdinosaur}; the
crop is the same on every frame of the window, and the window is flipped horizontally with
probability 0.5. Both encoders are trained to convergence on the 70 clips of the training half only,
at batch size 32. Discovery is scored on the four crops of each validation clip, each run over the
clip's 50 frames in one pass. Re-identification is scored on a fixed $189\times189$ crop of the
frame around the occluded car, resized to $378\times378$.

The dataset holds only 39 occlusion--reappearance events of cars, too few to train the walker on, so
we extend the pseudo-occlusions of Sec.~\ref{app:walker} to stand in for them entirely: a short stretch of
visible frames of a slot is declared hidden, and the target is the slot's own position on the frame after
it. On those frames the features of the patches the slot owns are replaced by features drawn
uniformly from the patches it does not own in the same frame, so the walker sees other content where
the object was, as under a real occluder, and no label enters the features, the anchor or the target.

\subsection{Datasets}
\label{app:datasets}

\paragraph{LA-CATER.}
Clips are 300 frames of $320\times240$ at 24\,fps.
9,300 / 3,327 / 1,371 training / validation / test clips under the static camera, and
9,299 / 2,152 / 1,371 under the moving camera of \citet{ram}.

\paragraph{The occlusions we score.}
A scorable occlusion is a maximal run of frames on which the target has no visible pixel, at least
five frames long, with three frames on each side on which the target covers at least 20\,px and is
owned by a resolved slot. For a target that stays in place, holding its last visible position is
already the right answer, and such occlusions make up 550 of the 695 under the static camera; we
therefore score every occlusion through which the target's amodal centre moves at least 10\,px in the
image (under the moving camera this includes camera motion), and cap the rest at 50, which still
penalise a model that moves a slot whose object stayed put. The 50 are drawn once from the ground truth
with a fixed seed, in proportion to the label and length (5--9, 10--29, $\geq$30 frames) of the
occlusions they are drawn from, and every row is scored on the same list. Table~\ref{tab:cases} counts them; Tables~\ref{tab:permanence} and~\ref{tab:same} are read on
them. The scored occlusions last 5 to 245 frames under the static camera and 5 to 240 under the moving
one, with medians of 3.2 and 1.8\,s.

\begin{table}[h]
\centering\scriptsize
\setlength{\tabcolsep}{6pt}
\begin{tabular}{@{}lcc@{}}
\toprule
& static camera & moving camera \\
\midrule
test clips read                                      & 1,200 & 1,200 \\
scorable occlusions                                  & 695   & 546 \\
\quad in clips                                       & 526   & 434 \\
\midrule
target moves $\geq$\,10\,px while hidden, all scored & 145   & 338 \\
target moves less, sampled                           & 50 of 550 & 50 of 208 \\
\midrule
occlusions scored                                    & \textbf{195} & \textbf{388} \\
\quad in clips                                       & 182   & 338 \\
\bottomrule
\end{tabular}
\caption{\textbf{From test clips to scored occlusions, LA-CATER.} The first 1,200 test clips of each
camera. Displacement is that of the target's amodal centre, between the first hidden frame and the
furthest point of the occlusion.
}
\label{tab:cases}
\end{table}

\paragraph{KITTI.}
We use the 21 labelled tracking videos of KITTI \citep{kitti}, split in time into training and
validation halves as in \citet{ram} and cut into clips of 50 frames. For occlusion analysis we
choose the car class and find its occlusions from the KITTI-MOTS masks \citep{mots}: runs of frames
on which a car's visible area stays below 15\,\% of its box and falls below 5\,\%, with the car
visible before and after. We collect 39 such occlusions of 29 cars from the 21 videos, lasting 0.1 to
2.2\,s.

\subsection{Baselines}
\label{app:baselines}

\paragraph{RAM.}
RAM's released evaluation post-processes its predictions on occluded frames, and its published LA-CATER
numbers include this step: for a target that was still before the occlusion, the prediction is replaced
by the last visible box or by a box placed on the nearest other detection. Neither rule carries a
position through the occlusion, and neither is available to a slot model, so for a fair comparison we
report RAM's raw output. RAM has no slots; it returns tracks, boxes that keep an identity across frames.
For SAME, the target's track on a frame is the one whose box overlaps the target's ground-truth box
most, and SAME asks whether that track has the same identity at the reappearance as before the
occlusion. The ground-truth box only selects which track is read and never enters a prediction.

\paragraph{Loci-Looped.}
Loci-Looped reads a background image of the scene beside the video, which its authors call the only
supervised input their model receives. We build it as their released loader does, from the ground-truth
instance masks: each pixel is its mean over the frames on which no object covers it. Under the moving
camera no single image fits a whole clip, so we rebuild it on every frame by inpainting the ground-truth
boxes of the visible objects, leaving the boxes of hidden objects untouched so that no hidden position
enters the input. Loci-Looped assigns pixels to slots by a mask rather than by attention, so the slot
that owns the target is the one whose mask, after depth competition, carries the largest mass on the
target's visible pixels; otherwise it is read as in Sec.~\ref{sec:exp:setup}, on its own steps, which
advance every second frame.

\paragraph{Checkpoints.}
\method{} and SlotContrast are read at step 100k, RandSF.Q at convergence (step 50k), and Loci-Looped at
convergence under the static camera and, under the moving one, at 170k, the last checkpoint before its
training collapsed at 171.5k without recovering.
\method{} and \emph{frozen} are read on
\ten{}'s states, SlotContrast and RandSF.Q on their own slots.

\subsection{Permanence metric}
\label{app:permmetric}

\begin{figure}[!htb]
\centering
\includegraphics[width=\textwidth]{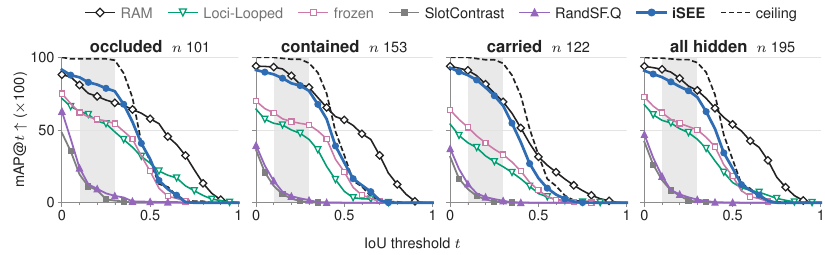}
\caption{\textbf{Permanence against the IoU threshold, LA-CATER static camera.}
mAP@$t$ $\uparrow$ ($\times100$). \emph{Grey band}: the thresholds 0.10 to 0.30 that
Table~\ref{tab:permanence} averages. \emph{Open markers}: rows given more than the video.
\emph{ceiling}: the share of occlusions on which \method{}'s box overlaps the ground-truth box with
IoU $>t$ on the last frame before the occlusion, where the box is on the target by construction.
The ceiling falls above $t=0.3$ because the box is cut from a $37\times37$ patch-grid mask, so
beyond it the metric reads the resolution of the grid, not the predicted position.}
\label{fig:map-thr}
\end{figure}

The metric is that of OPNet \citep{lacater}, averaged over IoU thresholds,
\begin{equation*}
\mathrm{mAP}@[0.1,0.3]=\frac{1}{5}\sum_{t\in\{0.10,0.15,\dots,0.30\}}\frac{1}{|\mathcal{O}|}
\sum_{o\in\mathcal{O}}\frac{1}{|F_o|}\sum_{f\in F_o}\mathbf{1}\big[\mathrm{IoU}(\hat b_f,b_f)>t\big],
\end{equation*}
with $\mathcal{O}$ the scored occlusions, $F_o$ the hidden frames of occlusion $o$ that carry the label
being scored, $\hat b_f$ the predicted box and $b_f$ the target's amodal ground-truth box. A slot model
predicts a position, not a box: as in \citet{ram}, its box has the size of the followed slot's grouping
mask on the last frame before the occlusion and is centred on the predicted position; the position is
read from the slot by a linear map fitted on visible frames of other clips. Loci-Looped is
scored on the box of its own amodal mask and RAM on its own box. The mask is cut from the $37\times37$
patch grid, so on the last frame before the occlusion, where the box is on the target by construction,
it overlaps the ground-truth box with IoU $>0.1$ on 99.5\,\% (static) and 96.9\,\% (moving) of the
occlusions but with IoU $>0.5$ on only 29.7\,\% and 17.8\,\%. Above 0.3 the metric measures the grid's
resolution rather than the predicted position (Fig.~\ref{fig:map-thr}), so we average the thresholds
0.10 to 0.30.

\subsection{Same-slot return (SAME)}
Re-identification asks one question per occlusion: once the object reappears, is it back in the slot
that held it before? SAME answers exactly this, once per occlusion, from two frames, the last before the
occlusion and the first after it. It is the identity switch of CLEAR MOT \citep{clearmot}, counted
across the occlusion only. Whole-clip identity metrics such as HOTA \citep{hota} and IDF1 \citep{idf1}
average over every frame of the clip. Most of those frames show the object in plain view, so an
identity lost at one occlusion moves them little. They also accept a prediction only above an overlap
threshold, which ties the identity score to the quality of the masks. Sec.~\ref{app:hota} reports HOTA.

A slot model makes the test direct:
\begin{itemize}
\item \emph{Every slot is already a track.} The recurrence carries each slot from frame to frame, so slot
$k$ on frame $t$ and slot $k$ on frame $t+1$ are the same slot, and no association step is needed.
\item \emph{No threshold.} Which slot holds an object is read from the slot attention itself.
\item \emph{No label while hidden.} The test needs the object's visible mask only at the two ends of the
occlusion, so no box and no label enters while the object is hidden.
\item \emph{No ground-truth identities.} The slot indices need not match any ground-truth identity,
because the test compares a model's slot with its own slot.
\end{itemize}

\paragraph{Definition.}
Let $A_t\in[0,1]^{K\times N}$ be the grouping attention on frame $t$, from the corrector's last
iteration and normalised over the $K$ slots. Let $\phi^j_{t,n}\in[0,1]$ be the fraction of patch $n$
covered by the visible (modal) mask $M^j_t$ of object $j$. The owner of $j$ on frame $t$ is the slot with
the largest attention mass on its visible pixels,
\begin{equation*}
\omega_t(j)=\arg\max_{k}\sum_{n=1}^{N}A_{t,k,n}\,\phi^j_{t,n},
\end{equation*}
defined when $M^j_t$ is not empty. Assigning an object to the slot whose mask overlaps it most is
standard in the slot literature. mBO gives each ground-truth mask the slot mask with the largest overlap
\citep{dinosaur,videosaur,slotcontrast}. The video forms of FG-ARI and mBO score each slot over the whole
clip, so a slot that switches objects is penalised \citep{savi,steve,videosaur,slotcontrast}.
\citet{tsa} pick the slot that tracks an object before an occlusion in the same way. Our owner is this
assignment on a single frame, with the slot's attention mass on the object's visible pixels in place of
the overlap of hard masks. The mass is absolute; normalising each slot's attention first would
favour a small slot lying inside the object. With $\bar M^j_t$ the amodal mask,
$v_t(j)=|M^j_t|/|\bar M^j_t|$ is the visible fraction. An occlusion $o$ of object $j_o$ is a maximal run
of frames $s_o\le t<e_o$ with $|M^{j_o}_t|=0$, lying inside the clip ($s_o>0$, $e_o<T$). Its two ends
are the last and the first frames around it on which at least half of the object is visible,
\begin{equation*}
t^-_o=\max\{t<s_o:\ v_t(j_o)\ge\tfrac12\},\qquad t^+_o=\min\{t\ge e_o:\ v_t(j_o)\ge\tfrac12\},
\end{equation*}
and over the set $\mathcal{O}$ of scored occlusions
\begin{equation*}
\mathrm{SAME}=\frac{1}{|\mathcal{O}|}\sum_{o\in\mathcal{O}}\mathbf{1}\big[\omega_{t^+_o}(j_o)=\omega_{t^-_o}(j_o)\big].
\end{equation*}
With the same rule at both ends, the owner is read only while the object is mostly in view. A frame
showing only a sliver of the object, where the owner is ambiguous, is never an end. An occlusion is
scored when both ends exist and $\omega_{t^-_o}$ is defined.

\paragraph{Scored occlusions.}
On LA-CATER, $\mathcal{O}$ is the target's occlusions of Sec.~\ref{app:datasets}, each matched to the
zero-visible run that overlaps it most. This scores 183 of the 195 static occlusions and 360 of the 388
moving ones. KITTI has no amodal mask, so the car's box stands in for it: $t^-_o$ is the last frame on
which the visible mask covers at least half the box, and $t^+_o$ the first on which the visible mask has
at least half the car's median visible area.

\paragraph{Models without slots.}
Loci-Looped's owner is read from its masks after depth competition, and RAM's from its tracks
(Sec.~\ref{app:baselines}).

\section{More experiments}

\subsection{Permanence and the length of the occlusion}

\emph{Frozen} loses a quarter of its permanence from the shortest occlusions to the longest and
Loci-Looped almost half, while \method{} and RAM lose less than 3 points (Table~\ref{tab:perm-length}).
\method{}'s lead over \emph{frozen}, paired per occlusion, therefore grows from $+13.3$ to $+28.3$ points.

\begin{table}[h]
\centering\scriptsize
\setlength{\tabcolsep}{5pt}
\begin{tabular}{@{}l cccc@{}}
\toprule
& \multicolumn{4}{c}{mAP@$[0.1,0.3]$ $\uparrow$, occlusion length (frames)} \\
\cmidrule(lr){2-5}
& 5--25 & 25--50 & 50--100 & $\geq$100 \\
& \scriptsize $n$ 50, median 7 & \scriptsize $n$ 24, median 38 & \scriptsize $n$ 50, median 77 & \scriptsize $n$ 71, median 150 \\
\midrule
\privrow \priv{RAM}         & \priv{86.0} & \priv{92.1} & \priv{82.6} & \priv{83.5} \\
frozen                     & 63.9 & 63.3 & 53.1 & 48.1 \\
Loci-Looped                & 69.3 & 55.7 & 49.6 & 37.1 \\
\midrule
SlotContrast               & 10.8 & 9.4  & 3.2  & 4.1 \\
RandSF.Q                   & 9.8  & 10.0 & 7.1  & 5.9 \\
\method{}                  & \textbf{77.1} & \textbf{91.7} & \textbf{81.8} & \textbf{76.4} \\
\bottomrule
\end{tabular}
\caption{\textbf{Permanence against the length of the occlusion, LA-CATER static camera.}
mAP@$[0.1,0.3]$ $\uparrow$ ($\times100$) over all hidden frames of the 195 occlusions, split by length
into four disjoint bins, with the rows and boxes of Table~\ref{tab:permanence}. \emph{frozen}: the last
visible position; RAM is given boxes, track identities and visibility labels, Loci-Looped a per-frame
background.
\method{} keeps its permanence as the occlusion lengthens, while \emph{frozen} and Loci-Looped fall.
Bold: best row that reads the video alone, which excludes \emph{frozen},
RAM and Loci-Looped; $n$ and median length in frames per bin in the header.
}
\label{tab:perm-length}
\end{table}

\subsection{Identity over whole clips}
\label{app:hota}

Table~\ref{tab:hota} scores identity over whole clips with HOTA \citep{hota}, computed with the official
TrackEval implementation.
\begin{itemize}
\item \emph{Tracks.} Each slot is a track; its detection on a frame is the region it wins in the argmax of the
decoder's masks, restricted to the foreground as in FG-ARI.
\item \emph{Ground truth.} The modal instance masks, matched by mask IoU, so an identity kept across an occlusion
is judged by the matches before and after it. \emph{Occluded objects}: objects with no visible pixel for at least
five consecutive frames that are visible again, the others joining the ignored background (our gate).
\emph{All objects}: every object of the same clips.
\item \ten{} raises association accuracy (AssA) on both encoders and costs one to three points of detection
accuracy (DetA). Without it, \method{} is below SlotContrast with the camera static and level with it with the
camera moving.
\item With the camera static, \method{} and SlotContrast\,+\,\ten{} are level on occluded objects and \method{}
leads in AssA over all objects; with the camera moving, \method{} leads in AssA on both object sets.
\end{itemize}

\begin{table}[h]
\centering\scriptsize
\setlength{\tabcolsep}{5pt}
\begin{tabular}{@{}l ccc ccc@{}}
\toprule
& \multicolumn{3}{c}{occluded objects} & \multicolumn{3}{c}{all objects} \\
\cmidrule(lr){2-4}\cmidrule(lr){5-7}
& HOTA $\uparrow$ & DetA $\uparrow$ & AssA $\uparrow$ & HOTA $\uparrow$ & DetA $\uparrow$ & AssA $\uparrow$ \\
\midrule
\multicolumn{7}{@{}l}{\emph{static camera} (150 clips; 224 occluded objects in 127 clips; 1,088 objects)} \\
SlotContrast             & 71.5 & 83.5 & 61.3 & 87.8 & 92.8 & 83.2 \\
\quad + \ten{}           & 81.3 & 82.6 & 80.0 & 89.5 & 92.0 & 87.1 \\
RandSF.Q                 & 42.3 & 61.8 & 29.0 & 35.1 & 48.1 & 25.9 \\
Loci-Looped              & 54.8 & 52.7 & 57.2 & 70.3 & 68.8 & 72.2 \\
\method{} without \ten{} & 68.0 & 82.3 & 56.2 & 85.4 & 92.1 & 79.3 \\
\method{}                & \textbf{82.1} & 81.2 & \textbf{82.9} & \textbf{90.4} & 91.0 & \textbf{89.9} \\
\midrule
\multicolumn{7}{@{}l}{\emph{moving camera} (150 clips; 226 occluded objects in 132 clips; 1,112 objects)} \\
SlotContrast             & 63.8 & 84.5 & 48.1 & 77.4 & 90.9 & 65.9 \\
\quad + \ten{}           & 68.6 & 82.5 & 57.1 & 80.0 & 89.7 & 71.4 \\
RandSF.Q                 & 42.0 & 56.4 & 31.4 & 26.6 & 28.8 & 25.0 \\
Loci-Looped              & 37.5 & 38.3 & 37.6 & 38.5 & 40.6 & 37.6 \\
\method{} without \ten{} & 63.6 & 82.1 & 49.2 & 76.0 & 88.1 & 65.6 \\
\method{}                & \textbf{75.3} & 79.1 & \textbf{71.7} & \textbf{81.3} & 85.0 & \textbf{77.8} \\
\bottomrule
\end{tabular}
\caption{\textbf{Identity over whole clips (TrackEval).} HOTA, DetA and AssA $\uparrow$ ($\times100$) on the first
150 test clips of each camera, protocol above. RandSF.Q: its decoder's masks (last-layer cross-attention,
head-averaged, softmax over slots). Loci-Looped: the argmax over its decoded masks with its background channel as
one more track, at every second frame, at the checkpoints of Sec.~\ref{app:baselines}.
}
\label{tab:hota}
\end{table}

\subsection{Separation of appearance and position}
Separation is the \emph{within-object} $R^2$ $\uparrow$ of a two-layer MLP that predicts an object's
centre from a block of its slot after the per-object mean is removed, so that the probe has to track
motion rather than recognise the object.

\begin{table}[t]
\centering\small
\setlength{\tabcolsep}{5pt}
\caption{\textbf{Position is carried by the position stream, MOVi-C.} Within-object
$R^2$ of an object's modal centre. Feature rows are each model's own frozen backbone pooled over the
object's cells, and are the floor for the code that reads them. Position is carried by the
4-dimensional position stream, and the appearance stream reads it below the floor.}
\label{tab:separation}
\begin{tabular}{@{}l c c@{}}
\toprule
code & dim & within-object $R^2$ \\
\midrule
DINOv2 \citep{dinov2}             & 384 & 0.833 \\
DINOv2 NoPE \citep{dinosaw}       & 384 & 0.552 \\
\midrule
SlotContrast slot                 & 64  & 0.597 \\
\method{} appearance $\mathbf{a}$ & 60  & 0.525 \\
\method{} position $\mathbf{p}$   & 4   & 0.727 \\
\bottomrule
\end{tabular}
\end{table}

In SlotContrast the position is carried by the whole 64-dimensional slot, the block its contrastive
term also acts on (Table~\ref{tab:separation}); in \method{} it is carried by the 4-dimensional position
stream, which reads more of it than the model's 384 input features, computed without a positional encoding.
The appearance stream's 0.525 is the position already present in the DINOv2 NoPE features it reads.
The probe is a read-out, so what it separates is not whether a code holds the position but how easily
that position can be used, which is what Sec.~\ref{sec:exp:dynamics} does with it: given each slot's
position delta as an action, the dynamics model on \method{}'s slots gains 47 points of headroom and the
one on SlotContrast's, given its whole slot delta, five.

\subsection{Contrastive loss on the whole slot}
\begin{table}[t]
\centering\small
\setlength{\tabcolsep}{4pt}
\caption{\textbf{Contrastive loss on the appearance stream or on the whole slot.} SAME, discovery,
permanence mAP and the within-object $R^2$ of the position stream for \method{} trained with the contrastive
loss on the appearance stream only (60 dimensions) or on the whole slot (64 dimensions), LA-CATER, static
camera.}
\label{tab:ssc-scope}
\begin{tabular}{@{}l cccccc@{}}
\toprule
contrastive loss on & SAME $\uparrow$ & FG-ARI $\uparrow$ & mBO $\uparrow$ & mAP all $\uparrow$ & mAP hidden $\uparrow$ & $R^2$ of $\mathbf{p}$ $\uparrow$ \\
\midrule
appearance stream (60) & 82.0 & 96.6 & 23.9 & 69.0 & 30.1 & 0.937 \\
whole slot (64)        & 83.1 & 94.7 & 28.5 & 25.6 & 13.0 & 0.693 \\
\bottomrule
\end{tabular}
\end{table}

With the contrastive loss on the whole slot, the other scores stay similar, but the position encoded in
the position stream degrades, and the permanence mAP, which is read from it, suffers as well
(Table~\ref{tab:ssc-scope}). The loss requires the same position to have a different code across the
sequence and across the batch, which a position code cannot satisfy.

\subsection{Appearance input of the position predictor}
\begin{table}[t]
\centering\small
\setlength{\tabcolsep}{4pt}
\caption{\textbf{Appearance input of the position predictor, LA-CATER, static camera.} Permanence
mAP@$[0.1,0.3]$ over all frames of the 195 occlusions, with \ten{} and without the walker.}
\label{tab:sga}
\begin{tabular}{@{}l c@{}}
\toprule
position predictor & mAP all $\uparrow$ \\
\midrule
with $\operatorname{sg}(\mathbf{a})$    & 69.0 \\
without $\operatorname{sg}(\mathbf{a})$ & 66.4 \\
\bottomrule
\end{tabular}
\end{table}

Without $\operatorname{sg}(\mathbf{a})$ the position predictor loses permanence mAP (Table~\ref{tab:sga}), as
expected, since how far an object moves depends on its shape.

\section{Dynamics model}
\label{app:dynmodel}
We train one OCVP-Par predictor \citep{ocvp} per encoder, with $L=4$ blocks of width 128 and 1.09M
parameters including the action path. It reads a 14-frame window of the encoder's frozen slots and
predicts the slots of the next frame; in a rollout, each prediction is appended to the window as its
newest frame. The predictor is trained on cached slots for 30k steps at seed 0, batch 128, learning
rate $2\times10^{-4}$ and dropout 0.2. The action of slot $k$ is its one-step position change
$\mathbf{p}^k_{t+1}-\mathbf{p}^k_t$, divided by its r.m.s.\ over the cache, mapped by a two-layer MLP
to a two-dimensional code and injected into every block by adaptive layer normalisation. The loss is a
squared error in which the four dimensions of $\mathbf{p}^k$ are weighted by $\lambda=88.7$ and the
sixty of $\mathbf{a}^k$ by 1; for SlotContrast, which has no position stream, it is unweighted.

\begin{figure}[!htb]
\centering
\includegraphics[width=0.709\linewidth]{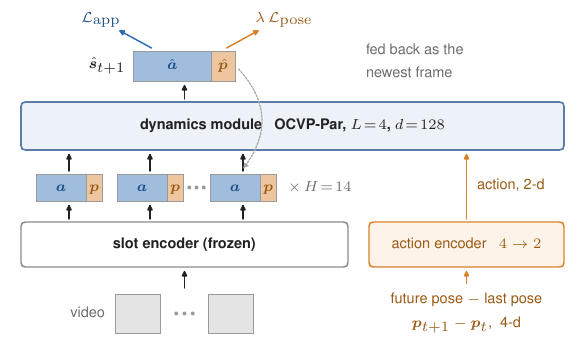}
\caption{\textbf{The action-conditioned dynamics model.} Frozen slots $[\mathbf{a}^k;\mathbf{p}^k]$
enter OCVP-Par \citep{ocvp}; per slot, the position change $\mathbf{p}^k_{t+1}-\mathbf{p}^k_t$ passes a
two-dimensional bottleneck and is injected into every block by adaptive layer normalisation. The
action reaches the predictor only through the position stream.}
\label{fig:dyn-arch}
\end{figure}

\clearpage
\section{Qualitative examples}
\label{app:qual}

\subsection{Object discovery}
\begin{figure}[h!]
\centering
\includegraphics[width=\textwidth]{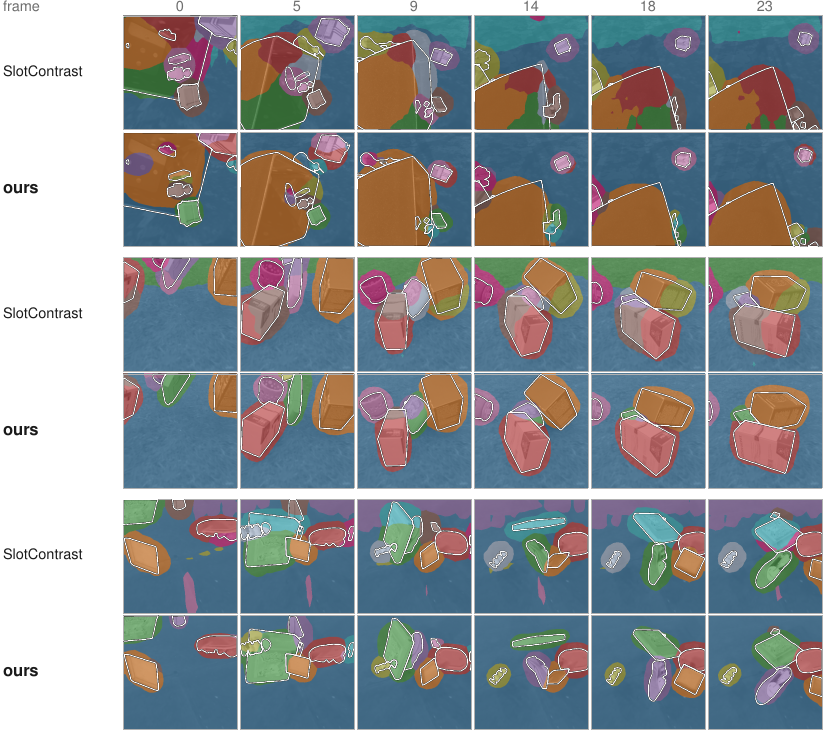}
\caption{\textbf{Object discovery on MOVi-C.}}
\label{fig:qual-disc-movic}
\end{figure}

\begin{figure}[h!]
\centering
\includegraphics[width=\textwidth]{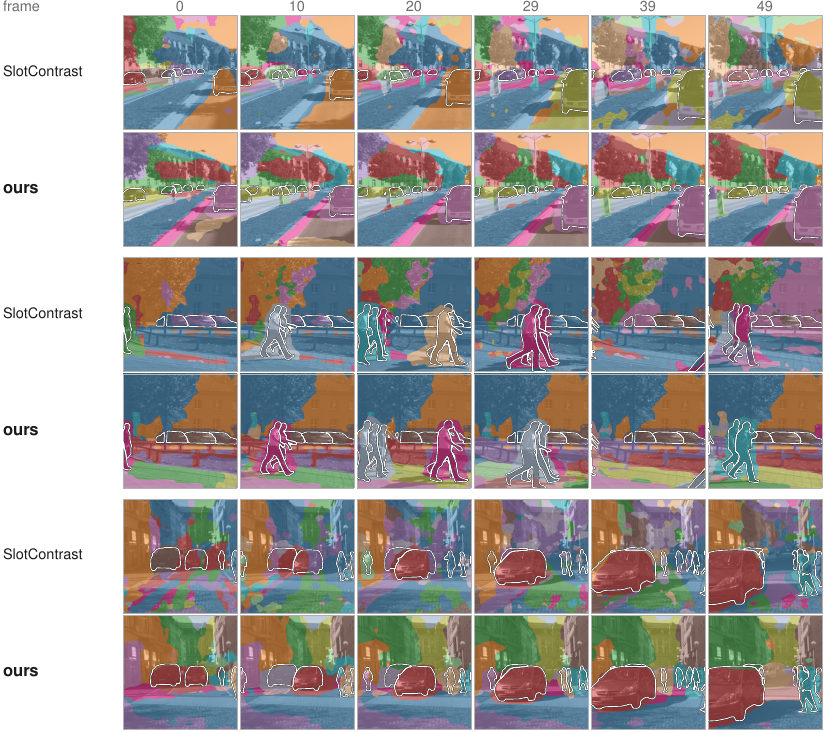}
\caption{\textbf{Object discovery on KITTI.}}
\label{fig:qual-disc-kitti}
\end{figure}

Figures~\ref{fig:qual-disc-movic} and~\ref{fig:qual-disc-kitti} draw the decoder-mask argmax of SlotContrast and
\method{} on MOVi-C and on KITTI; Table~\ref{tab:discovery} is the population these two clips are drawn from.

\clearpage

\clearpage

\subsection{Re-identification on LA-CATER}
\begin{figure}[t]
\centering
\includegraphics[width=\textwidth]{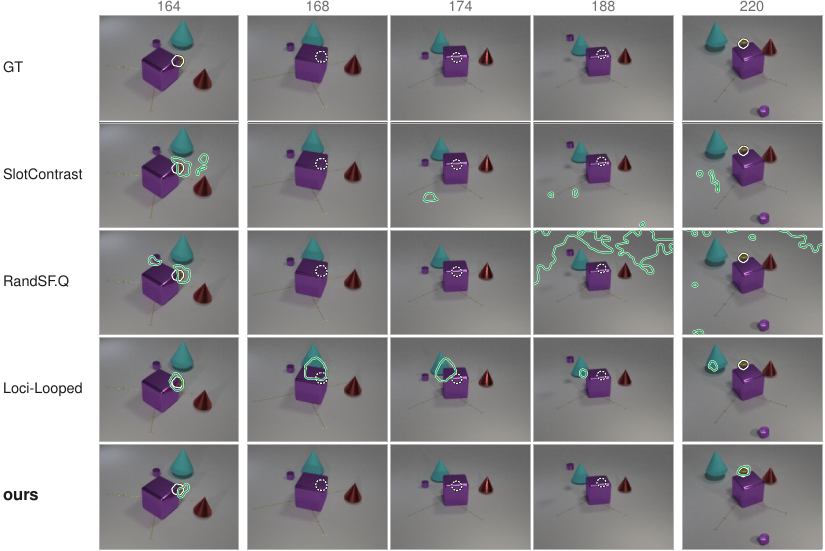}
\caption{\textbf{Re-identification: one occlusion in a LA-CATER scene, the target hidden behind another object and reappearing.}
Columns are frames, the first the last one on which the target is visible and the last the frame
Table~\ref{tab:same} scores, in one crop of the frame that is the same throughout. \emph{GT}: the target's true
(amodal) mask, in white on every row, dashed on the frames where none of it is visible. Green: the mask of the slot
that owns the target in the first column, empty where that slot claims nothing in view; \method{}'s claims nothing
anywhere while the target is hidden, because \ten{} holds it. Only \method{} has that slot back on the target when
it reappears.}
\label{fig:qual-reid}
\end{figure}

Figure~\ref{fig:qual-reid} follows one target through an occlusion and back out, and asks whether the slot that
held it before still holds it after; Table~\ref{tab:same} is the population.

\begin{figure}[t]
\centering
\includegraphics[width=\textwidth]{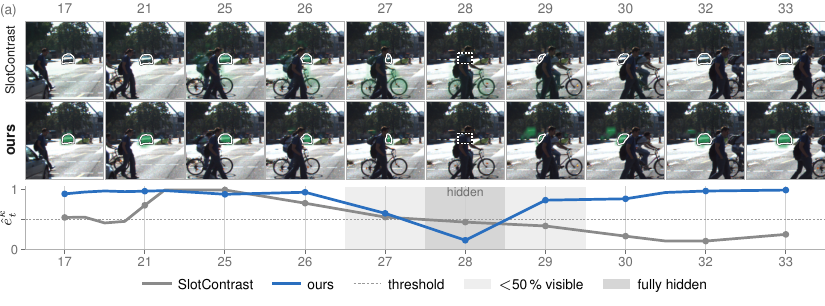}
\caption{\textbf{A qualitative example on KITTI.}
Columns are frames of one tile, unevenly spaced; the whole $2\times$ tile the model sees. \emph{GT}: the car's
visible mask in white, its KITTI box dashed where none of it is visible. Green: the slot on the car before the
occlusion. \method{}'s evidence falls while the car is hidden and its slot is back on the car when it
reappears; SlotContrast's slot moves onto the occluding cyclist.}
\label{fig:kitti-reid-supp}
\end{figure}

\clearpage

\subsection{Object permanence on LA-CATER}
\begin{figure}[h!]
\centering
\includegraphics[width=\textwidth]{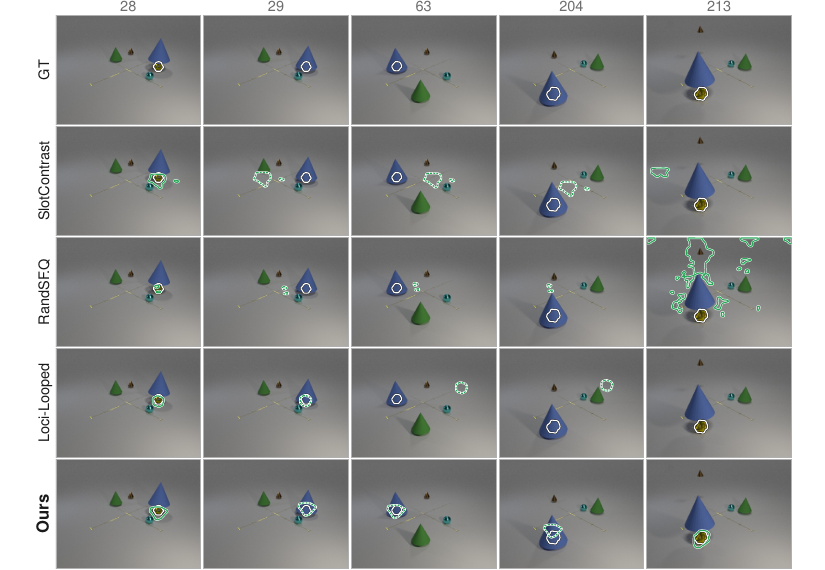}
\caption{\textbf{Qualitative comparative example on LA-CATER, static camera}, against SlotContrast, RandSF.Q and
Loci-Looped.}
\label{fig:qual-perm-static2}
\end{figure}

\begin{figure}[h!]
\centering
\includegraphics[width=\textwidth]{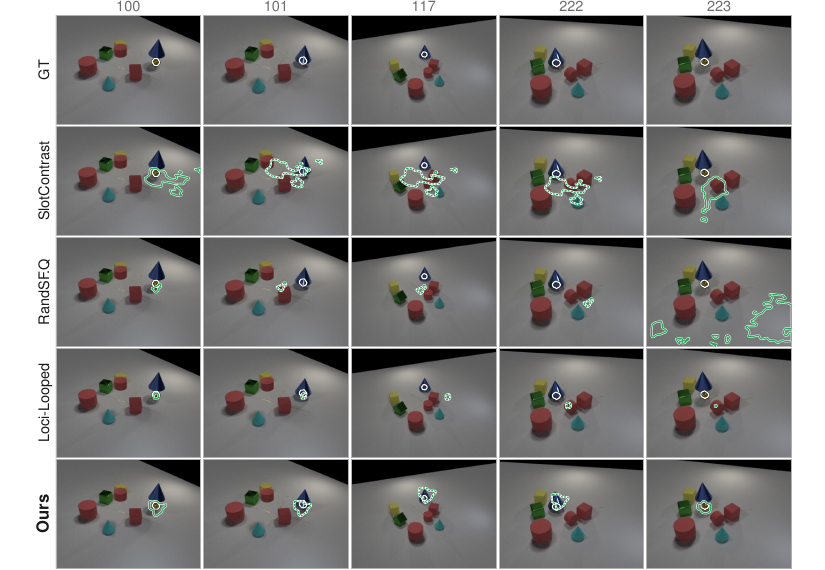}
\caption{\textbf{Qualitative comparative example on LA-CATER, moving camera}, against SlotContrast, RandSF.Q and
Loci-Looped.}
\label{fig:qual-perm-moving}
\end{figure}

Figures~\ref{fig:qual-perm-static2} and~\ref{fig:qual-perm-moving} draw one occlusion each in the form of
Figure~\ref{fig:qual-perm}, in a second static-camera scene and on the moving-camera split. Each is the occlusion
with the largest margin of \method{} over the other label-free rows of Table~\ref{tab:permanence}, so both are best
cases.

\clearpage

\clearpage

\subsection{Failure cases}
\subsubsection{Re-identification}
\begin{figure}[t]
\centering
\includegraphics[width=\textwidth]{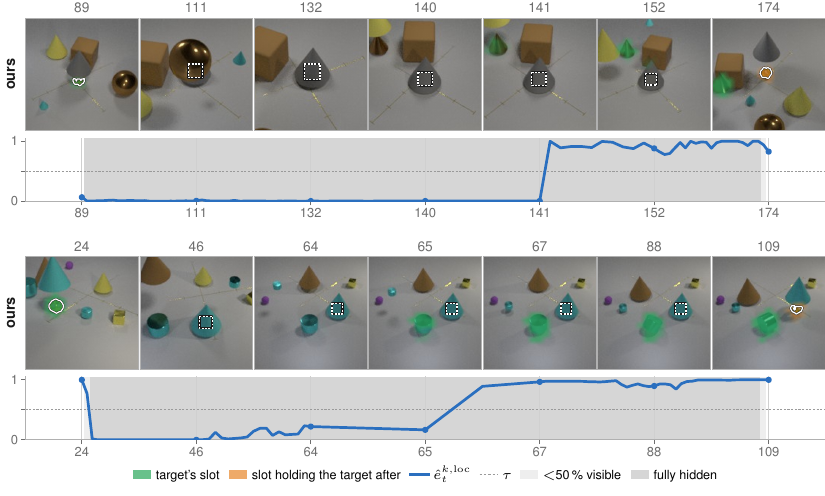}
\caption{\textbf{\ten{} releases the slot while the target is still hidden.}
Columns are frames of one clip, unevenly spaced, in one crop held fixed through the sequence; the first is the
last frame on which the target is at least half visible. \emph{GT}: the target's visible mask in white, its amodal
box dashed where none of it is visible. Green: the slot on the target before the occlusion; orange: the slot holding
the target after.}
\label{fig:lacater-fail}
\end{figure}

\begin{figure}[t]
\centering
\includegraphics[width=\textwidth]{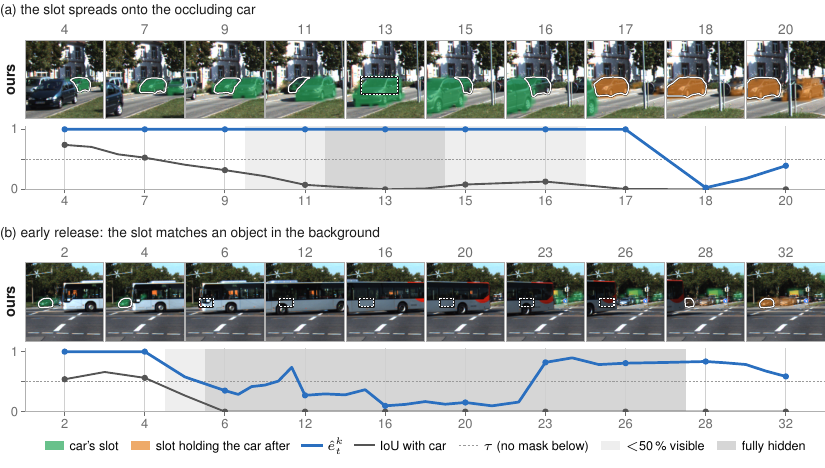}
\caption{\textbf{Two ways \method{} loses a car through a KITTI occlusion.}
Columns are frames of one tile, unevenly spaced. \emph{GT}: the car's visible mask in white, its KITTI box dashed
where none of it is visible. Green: the slot on the car before the occlusion.}
\label{fig:kitti-fail}
\end{figure}

Figure~\ref{fig:lacater-fail} draws two moving-camera LA-CATER occlusions on which \ten{} releases the slot while
the target is still hidden, and the slot takes another object. Figure~\ref{fig:kitti-fail} draws two ways
\method{} loses a car through a KITTI occlusion.

\clearpage

\subsubsection{Object permanence}
\begin{figure}[h!]
\centering
\includegraphics[width=\textwidth]{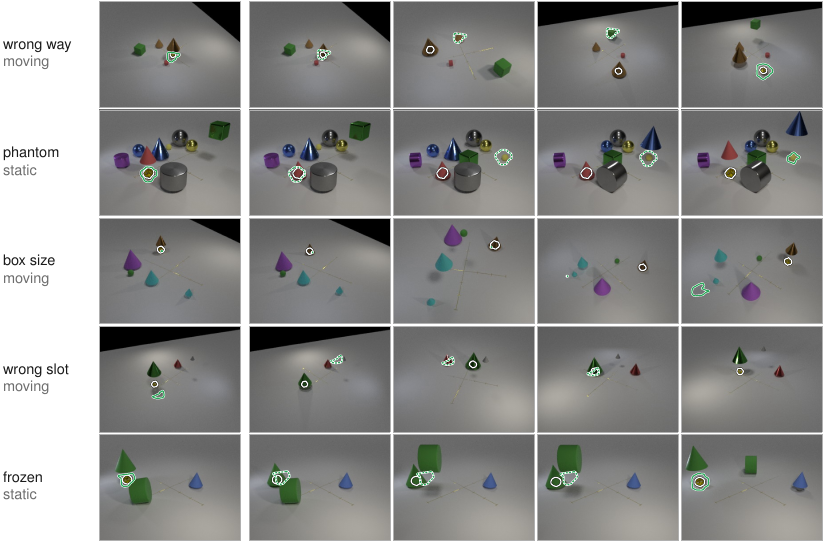}
\caption{\textbf{Five ways \method{} loses the target through a LA-CATER occlusion.}
\emph{GT}: the target's true (amodal) mask, in white. Green: the slot \method{} follows, \emph{solid} where it is
that slot's own mask and \emph{dashed} where that outline is moved onto the predicted position; no green where the
slot owns no pixel.}
\label{fig:perm-fail}
\end{figure}

Figure~\ref{fig:perm-fail} draws one LA-CATER occlusion for each of five ways \method{}'s predicted position
fails: it moves to the wrong place (\emph{wrong way}), it moves while the target stays still (\emph{phantom}), the
slot's box is too small to reach $\mathrm{IoU}>0.1$ (\emph{box size}), the slot is not on the target when it goes
in (\emph{wrong slot}), or it stays still while the target moves (\emph{frozen}). Each row is the median occlusion
of its mode, not the worst.

\clearpage

\end{document}